\documentclass[journal]{IEEEtran}

\usepackage[colorlinks,urlcolor=blue,linkcolor=blue,citecolor=blue]{hyperref}
\usepackage{amsmath}
\usepackage{color,array}
\usepackage{wrapfig}
\usepackage{bm}
\usepackage{extarrows}
\usepackage{amsmath, amssymb, amsfonts} 
\usepackage{algpseudocode}
\usepackage{graphicx}
\usepackage[ruled]{algorithm2e}
\usepackage{algorithmicx}
\usepackage{algpseudocode}

\usepackage{subfigure}

\usepackage{CJK}
\usepackage[top=2cm, bottom=2cm, left=2cm, right=2cm]{geometry}

\usepackage{amsmath}
\usepackage{indentfirst}
\usepackage{wrapfig}

\usepackage{multicol}
\usepackage{afterpage}
\begin{document}

\title{Domain Similarity-Perceived Label Assignment for Domain Generalized Underwater Object Detection}

\author{Xisheng Li, Wei Li$^*$, Pinhao Song, Mingjun Zhang, and Jie Zhou 
\thanks{Xisheng Li. School of Artificial Intelligence and Computer Science, Jiangnan University, Jiangsu, P. R. China Email: 6213113079@stu.jiangnan.edu.cn}
\thanks{Wei Li*. Corresponding author. School of Artificial Intelligence and Computer Science, Jiangnan University, Jiangsu, P. R. China. Email: cs\_weili@jiangnan.edu.cn}
\thanks{Pinhao Song are with Robotics Research Group, the Department of Mechanical Engineering, KU Leuven, Belgium. Email: pinhao.song@kuleuven.be}
\thanks{Mingjun Zhang. School of Artificial Intelligence and Computer Science  Jiangnan University, Jiangsu, P. R. China. Email: mingjunzhang@stu.jiangnan.edu.cn}
\thanks{Jie Zhou. School of Artificial Intelligence and Computer Science  Jiangnan University, Jiangsu, P. R. China. Email: 7223115006@stu.jiangnan.edu.cn}
}

\markboth{}
{First A. Author \MakeLowercase{\textit{et al.}}: Bare Demo of IEEEtai.cls for IEEE Journals of IEEE Transactions on Artificial Intelligence}

\maketitle

\begin{abstract}
The inherent characteristics and light fluctuations of water bodies give rise to the huge difference between different layers and regions in underwater environments. When the test set is collected in a different marine area from the training
set, the issue of domain shift emerges, significantly compromising the model’s ability to generalize. The Domain
Adversarial Learning (DAL) training strategy has been previously utilized to tackle such challenges. However, DAL heavily depends on manually one-hot domain labels, which implies no difference among the samples in the same domain. Such an assumption results in the instability of DAL. This paper introduces the concept of Domain Similarity-Perceived Label Assignment (DSP). The domain label for each image is regarded as its similarity to the specified domains. Through domain-specific
data augmentation techniques, we achieved state-of-the-art
results on the underwater cross-domain object detection
benchmark S-UODAC2020. Furthermore, we validated the
effectiveness of our method in the Cityscapes
dataset.
\end{abstract}


\section{INTRODUCTION}
\IEEEPARstart {O}{bject} detection is a critical task in computer vision that aims to automatically identify specific objects in images or videos and precisely locate them. It has significant applications in intelligent surveillance, autonomous driving, and robot navigation. Traditional object detection algorithms \cite{liu2016ssd,girshick2015fast,cai2018cascade,girshick2014rich}, assume that the training and testing datasets are sampled from the same distribution, sharing similarities in image features, scene settings, and data collection methods. However, underwater scenarios present unique challenges. The testing dataset often deviates from the training dataset due to variations in lighting conditions, color distortion, light attenuation (coastal, deep oceanic, murky waters), and camera settings \cite{uplavikar2019all,liu2020wqt}. 

Domain generalization (DG) is a concept in machine learning that involves training a model on data from multiple different but related domains so that it can perform well on unseen domains. The actual application in the real underwater environment matches the DG definition due to light fluctuation and attenuation in different water bodies. However, underwater cross-domain scenarios have received comparatively limited attention. In the scarce studies on cross-domain underwater object detection, researchers have incorporated domain generalization training strategies, leading to significant improvements in cross-domain scenarios \cite{chen2023achieving}. 

Domain Adversarial Learning (DAL) as proposed by \cite{ganin2015unsupervised}, employs domain adversarial learning to align features across underwater cross-domain scenarios, which is widely used in DG. This approach significantly improves the generalization capability of detection models in underwater environments, resulting in enhanced performance. However, DAL faces certain challenges in its application to cross-domain object detection underwater. (i) The existing DG dataset utilizes domain labels annotated manually for applying DAL. The real world includes a mixture of domains that are difficult to explicitly annotate. (ii) Even with a significant investment of human resources in annotating the dataset with discrete domain labels, obtaining favorable detection outcomes proves challenging. This issue arises from a phenomenon emphasized in \cite{zhang2023free}, wherein the high similarity between two domains, when artificially assigned distinct domain labels, can negatively impact the training stability of DAL. Specifically, the backbone may extract highly similar features from these two domains, leading to the domain discriminator overfitting to these inaccurately labeled examples. This, in turn, compromises the model's generalization ability \cite{thanh2019improving}. (iii) Additionally, Domain Data Augmentation (DDA) is a commonly employed technique in DG problems. When we expand the number of domains, manually labeling domain labels becomes impractical. Therefore, existing DAL methods fail to benefit from domain data augmentation.

To address the aforementioned issue, we propose Domain Similarity-Perceived Label Assignment (DSP), which eliminates the need for manual annotations. The central concept of our approach is to perceive a domain as a blend of similarities with various other domains. Each domain is regarded as a sample within a continuous space, enabling the direct generation of distinct pseudo-domain labels for individual images. Inspired by Farthest Point Sampling \cite{qi2017pointnet++}, we leverage Farthest Feature Sampling (FFS) to autonomously curate a set of base domains from the source domain without requiring input from the dataset. Subsequently, the domain label for an image is determined by its similarity to this set of base domains. The objective of designing the DSP module is to train a domain classifier capable of discerning among these base domains. Performing inference using a trained domain classifier, pseudo domain labels can be generated and presented in a soft label format, as opposed to discrete labels. This labeling approach enhances the stability of the DAL training process.

It can be concluded that a detector trained across a wide range of domains demonstrates domain invariance. Therefore, increasing sampling across the domain distribution contributes to enhanced robustness against domain shifts. \cite{chen2023achieving} Consequently, Domain Data Augmentation (DDA) has emerged as a crucial technique in DG. However, since DDA generates images belonging to various domains, annotating domain labels becomes impractical. By training DSP, we can uncover similarities within the newly generated domains. Leveraging the similarity among a few domains with maximum style differences allows us to effectively represent the remaining domains. We employ the Spurious Correlations Generator (SCG) \cite{xu2023multi} to generate a significantly larger number of domains compared to the original set, and then apply our DSP to label these domains. By combining SCG, DSP, and DAL, we achieved state-of-the-art results in underwater cross-domain object detection benchmark S-UODAC2020. Furthermore, we validated the effectiveness of our approach on the more general scenario of Cityscapes. 
\begin{itemize}

\item We introduce Domain Similarity-Perceived Label Assignment (DSP), which provides pseudo-domain labels for Domain Adversarial Learning (DAL). This not only economizes human resources but also mitigates the over-confidence associated with traditional one-hot labels.

\item We integrate DSP with the domain data augmentation SCG and train the model in the DAL framework.

\item  We achieve state-of-the-art results in underwater cross-domain object detection benchmark S-UODAC2020 and demonstrate versatility on the Cityscapes dataset.

\end{itemize}

\section{LITERATURE REVIEW}
\subsection{Object detection}
Driven by the swift evolution of neural networks, object detection has emerged as a fundamental task in computer vision. Its primary objective is to localize and classify distinct instances within various images. Modern object detection can be roughly categorized into two-stage \cite{cai2018cascade,ren2015faster,girshick2015fast,he2017mask,wang2017fast} and one-stage detectors \cite{redmon2016you,redmon2017yolo9000,redmon2018yolov3,liu2016ssd}. Two-stage methods have two main steps. First, they generate candidate regions likely to contain objects in the ``Region Proposal'' stage. Then, in the ``Object Classification and Localization'' stage, they use classifiers to determine objects within these regions and refine the object's position using regression. One-stage methods detect objects in a single step. They predict across multiple scales by processing network layers directly. This makes them simpler and faster, suitable for real-time applications, albeit potentially sacrificing some accuracy. However, traditional object detection methods struggle with domain shift challenges arising from differences in data distributions, scene assumptions, and label variations, necessitating the development of domain generalization object detection to enhance cross-domain adaptability.

\subsection{Underwater Object Detection}
Lately, researchers from across the globe have shown significant interest in the field of underwater object detection. A particular category of methodologies centers around the application of data augmentation techniques \cite{huang2019faster,liu2020towards}, with ROIMIX \cite{lin2020roimix} being a prominent example. It leverages the strategic implementation of mixup at the Region of Interest (RoI) level. Other methods like SWIPENET \cite{chen2022swipenet} maximize the benefits inherent in high-resolution and semantically enriched hyperfeature maps to enhance the detection accuracy of smaller objects. Attention mechanisms \cite{liang2022excavating} and feature pyramid strategies \cite{zhao2021composited} have also demonstrated advancements in the extraction of features in underwater conditions. Hard example mining \cite{song2023boosting} is also beneficial for detecting vague objects in the underwater environment.
The aforementioned methods are all based on the assumption that the training and testing datasets adhere to the same data distribution, without taking into consideration domain shift. To the best of our knowledge, there are only a few existing works that address this particular issue. First, DG-YOLO \cite{liu2020towards} takes the lead in addressing the underwater domain shift issue by pioneering the concept of domain-generalized underwater object detection. It employs WQT \cite{yoo2019photorealistic} to broaden the dataset's water conditions. Then, it uses adversarial learning and IRM \cite{arjovsky2019invariant} techniques to help the model generalize better. DMCL \cite{chen2023achieving} set up a benchmark S-UODAC2020 to evaluate how well models can adapt and proposed a domain mixup and contrastive Learning paradigm. We follow the domain generalization benchmark to address the issue of underwater domain shift.

\subsection{Domain Adaptation and Generalization}
In practical applications, data distributions often vary across different environments due to changing circumstances, selection bias, or time shifts. When the distributions of training and test data are indeed different, it can lead to a significant degradation in model performance \cite{szegedy2013intriguing,recht2019imagenet,torralba2011unbiased}. Domain Adaptation (DA) involves using data from the target domain directly for adaptation, aiming to improve model performance on that specific domain. Many approaches have been developed to tackle domain adaptation within recognition tasks. In unsupervised domain adaptation, various techniques are utilized to align the distribution in the pixel space. \cite{bousmalis2017unsupervised} or feature space \cite{ganin2015unsupervised} by adversarial learning. Numerous methods for Domain-Adaptive Object Detection (DAOD) have been introduced \cite{cai2019exploring,chen2018domain,hsu2020every}, mainly divided into adversarial-based methods and reconstruction-based methods.

Domain generalization (DG) is introduced to train a model that remains effective when faced with new and unfamiliar domains, using one or multiple source domains for its training. Existing methods can be categorized into three different types. First, diversifying source data within the domain via domain augmentation \cite{shankar2018generalizing,volpi2018generalizing,zhou2020deep}. Second, aligning features of the source domains \cite{motiian2017unified,li2018deep}. Third, designing customized modules \cite{choi2021robustnet,wu2022single}. Furthermore, there are several highly innovative approaches, like the self-supervised training method JiGEN \cite{carlucci2019domain}, which improves the generalizability by solving the jigsaw puzzle problem. 

DAL is a popular technique to align source domains in DG \cite{li2018domain}. However, using one-hot labels poses a challenge as it leads to over-confidence issues due to the inherent correlations among source domains. \cite{zhang2023free} attempts to address this concern by introducing Environment Label Smoothing (ELS). Nevertheless, ELS involves evenly distributing the high-confidence class probabilities to other classes, failing to adequately reflect the inter-domain relationships. In contrast, the pseudo labels generated by DSP take the form of soft labels, representing the similarity to specific domains. It ensures that each image's domain is distinctly characterized, aligning more closely with the training objectives of DAL.

\section{METHODS}
\subsection{Overview}

\begin{figure*}[htp]
    \centering
    \includegraphics[scale=0.75]{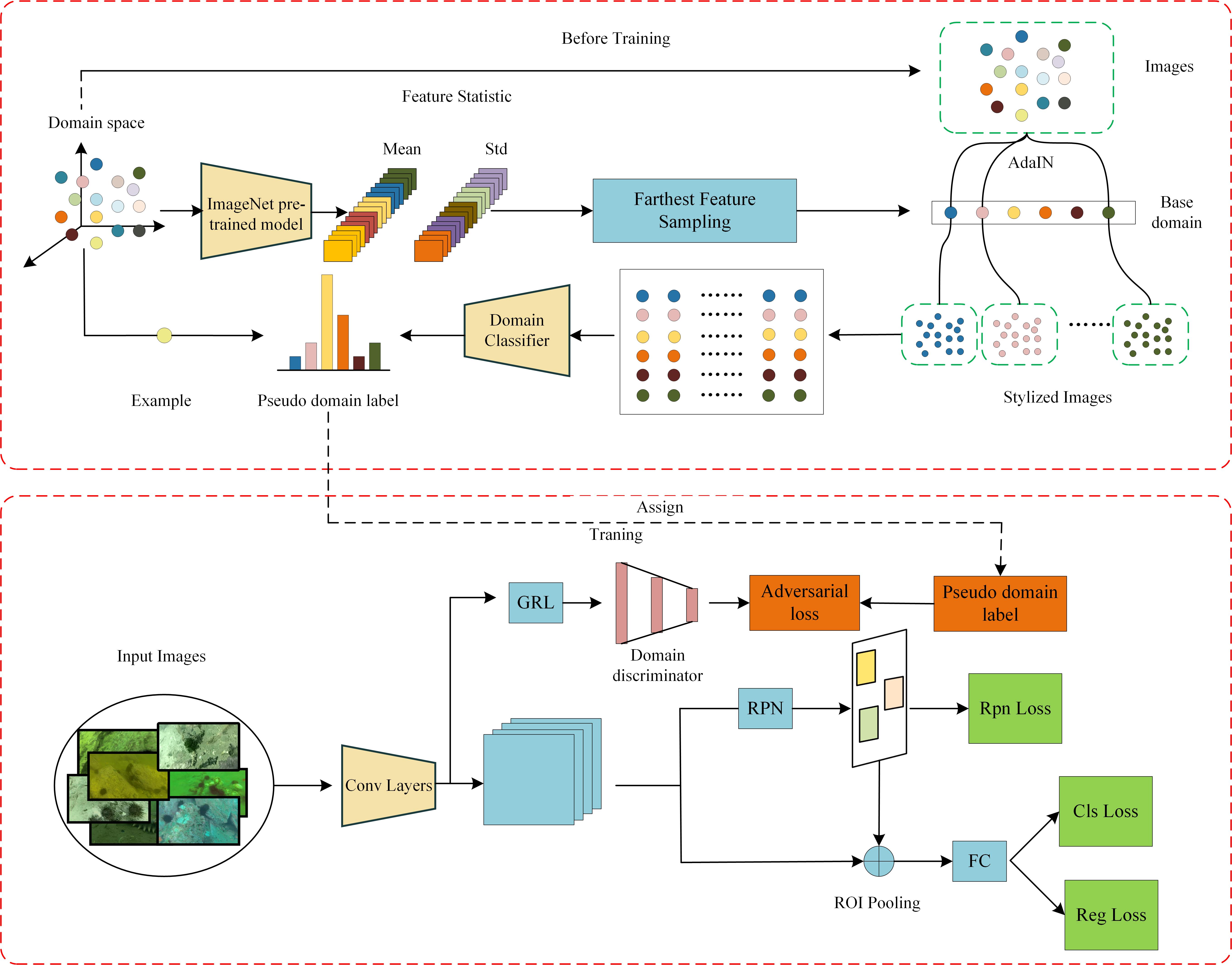}
    \caption{DSP is a preprocessing module used before model training. It utilizes a pre-trained model from ImageNet to extract low-level semantic information from images and aggregate their feature statistics. These statistics are then stacked together, and base domains are selected from them using the Farthest Feature Sampling method. Subsequently, each base domain's image quantity is augmented using Adaptive Instance Normalization (AdaIN). These augmented images are fed into the Domain Classifier. Finally, the Input Images are passed through the Domain Classifier for inference, yielding domain labels.}
    \label{DSP}
\end{figure*}

Following the usual terms for domain generalization, we define $\{\bm{x}_i\}^N_{i=1} \in \mathcal{D}_\mathcal{S}$ where $\bm{x}_i$ are the samples, $\mathcal{D}_\mathcal{S}$ is the source domain. 
In the typical DG problem, the source domain can be manually divided into several subsets to obtain domain labels. However, this assumption is not practical in the actual application because the real-world data is a mixture of domains that is difficult to annotate.
Therefore, we introduce Domain Similarity-Perceived Label Assignment (DSP) to construct $K$ most dissimilar source domains within the dataset $\mathcal{D}_\mathcal{S} = \{\mathcal{D}_{\mathcal{S}_1}, ..., \mathcal{D}_{\mathcal{S}_K}\}$ and use them as a reference to generate pseudo-domain labels for each image. Consequently, DAL can be trained even without the domain labels in the training dataset.

\subsection{Domain similarity-perceived label assignment (DSP)}
We aim to obtain pseudo domain labels for every image $\bar{\bm{y}}_i= f(\bm{x}_i) \in \mathbb{R}^K$. $f(\cdot)$ is the proposed label assignment function. As illustrated in Fig \ref{DSP}, DSP utilizes Farthest Feature Sampling (FFS) to obtain this set of base domains. AdaIN is employed to expand the number of base domains for training the domain classifier  Inference with this domain classifier allows us to obtain a pseudo-domain label for each image.

\subsubsection{Farthest Feature Sampling (FFS)}
To construct a set of the base domains, we aim to find images that are most different from each other in style. Inspired by Farthest Point Sampling which is used to downsample the point cloud, we propose Farthest Feature Sampling (FFS). The entire process can be found in Algorithm \ref{algorithm1}. The goal of FFS is to select K images from the dataset that exhibit the farthest style distances from each other, as $\{\bar{\bm{x}}_{k}\}_{k=1}^{K} = \mathop{FFS}(\{\bm{x}_{i}\}_{i=1}^{N})$. Various works \cite{gatys2016image,li2016combining,li2017demystifying} suggest that The convolutional feature statistics encode the style in an image, which can be used to calculate the style distance. 
In detail, we first passed all data through the pre-trained model, took the low-level features, then calculated their mean and standard deviation $\mathcal{F} = $$\{({\mu(\phi(x_1),\sigma(\phi(x_1)),....
,(\mu(\phi(x_N),\sigma(\phi(x_N))}\}$, $\phi$ denote the low-level layer in backbone. In this way, each image can correspond to a set of feature statistics. We aim to use these statistical features to find the $K$ domains that best represent the training set. Secondly, select a random image as the starting point, denoted as $\hat{d}_0$, and add it to the set $C$, which means including already selected domains for storage. We denote $K$ as the intended number of base domains. Calculate the style distances from $\mathcal{F}$ to the selected set $C$ and add the image with the farthest distance to $C$. After $K$ steps, we have K images as base domains.

\begin{algorithm*}
  \SetAlgoLined
  \textbf{Given:}$\{\bm{x}_i\}^N_{i=1} \in \mathcal{D}_\mathcal{S}$: all images in source domain, $\phi$ is shallow layer feature extractor.

  \KwResult{Obtain $K$ images with the farthest style distances.}

  \textbf{Initialization:}\\ {1. Compute feature statistics \\
  $\mu_n = \mu(\phi(\bm{x}_n)), \sigma_n = \sigma(\phi(\bm{x}_n))$ \\
  $\mathcal{F} = $$\{{(\mu_1, \sigma_1),..., (\mu_n,\sigma_n),..., (\mu_N,\sigma_N)}\}$}\\
  {2. $S \in \mathbb{R}^{N}$ stores distances from selected set $C$ to set $\mathcal{D}_\mathcal{S}$, which is initialized to $\infty$.} \\
  {3. Select a random image $\tilde{\bm{x}} \in 
 \mathcal{D}_\mathcal{S}$, $\tilde{\bm{x}} \rightarrow C$}\\
  
\For{$i \gets 0$ to $N-1$}{
       {Calculate style distances $d$ between $\mathcal{F}$ and $C$ using} \\
     \For{$j \gets 0$ to $N-1$}{
       \hspace{0.5em}$d_j = \sum\limits_{\bm{x}_l \in C} \|\mathcal{F}_j - (\mu(\phi(\bm{x}_l)), \sigma(\phi(\bm{x}_l))\|^2$ \\
      $S_j = d_j$ if $S_j > d_j$ 
       }

      $\bm{x}_k \rightarrow C$, where $k = \arg\max_j(S_j)$
    }
  \caption{Farthest Feature Sampling}
  \label{algorithm1}
\end{algorithm*}

\subsubsection{Real-time Arbitrary Style Transfer (AdaIN)}
The pseudo domain label for each image is determined by its similarity to the set of $K$ images. Hence, we need to train a classifier capable of distinguishing these $K$ base domains. We have only one image for each base domain, which is insufficient for training a domain classifier. Therefore, we need to transform our existing images through style transfer to match the style of the base domain images. The goal of style transfer is to create a new image that is based on the content of one image but rendered in the style of another. AdaIN combines Instance Normalization with style transfer and it allows us to adaptively adjust the features of an input image based on the style of a reference image. It computes the mean and variance of the input features and then rescales these features using the mean and variance of the reference image to match its style. In this way, the input image will be stylized into the style of the reference image. After obtaining the image of $K$ base domains using FFS, we intend to train a domain classifier by transforming the images of the dataset into representations of these styles using AdaIN, as:
\begin{align}
& \tilde{\bm{f}_i^k} = \sigma(\phi(\bar{\bm{x}}_k))\frac{\phi(\bm{x}_i)-\mu(\phi(\bm{x}_i))}{\sigma(\phi(\bm{x}_i))}+\mu(\phi(\bar{\bm{x}}_k)),\\
& \tilde{\bm{x}}^k_{i} = \phi^{-1}(\tilde{\bm{f}_i^k})
\end{align}
where $\phi(\cdot)$ and $\phi^{-1}(\cdot)$ are the ImageNet pre-trained model and the inverse decoding model, respectively. $\tilde{\bm{x}}^k_{i}$ is the stylized $\bm{x}_i$ by style $\bar{\bm{x}}_k$.
With the stylized dataset, the domain classifier can be trained with the goal as:
\begin{align}
& f = \mathop{argmin}_{f} \sum_{i=1}^{N} \sum_{k=1}^{K} \mathop{log} \tilde{\bm{y}}^k_i(k),\\
 &   \tilde{\bm{y}}^k_i= f(\tilde{\bm{x}}^k_{i}) \in \mathbb{R}^K,
\end{align}
With the trained domain classifier (label assignment function) $f(\cdot)$, a unique domain label for each image can be obtained for DAL.

\subsection{Domain Adversarial Learning (DAL)}
DAL is a conventional approach for capturing shared characteristics among diverse domains. It achieves this by maximizing the cost of the domain discriminator. 
Our approach differs from traditional DAL in that we do not aim to confuse artificially defined discrete domains. Instead, we aim to perturb the entire domain space represented by all images in the training set. The domain adversarial loss can be written as:

\begin{equation}
 \mathcal{L}_{dal} = \mathop{max}_{h} \sum_{i=1}^{N} CE(h(g(\bm{x}_i)), f(\bm{x}_i))
\end{equation}
where $N$ denotes the total number of images, $g(\cdot)$ for backbone, $h(\cdot)$ for domain discriminator. The domain discriminator aims to maximize the loss with pseudo-domain labels. while the backbone aims to confuse the domain discriminator. We add this loss in the object detection task, and obtain the total loss as:
\begin{equation}
\mathcal{L}_{\text{tot}} = \mathcal{L}_{\text{rpn}} + \mathcal{L}_{\text{cls}} + \mathcal{L}_{\text{loc}} + \mathcal{L}_{\text{dal}}
\label{3}
\end{equation}
where $\mathcal{L}_{\text{rpn}}$ denoting the Region Proposal Network (RPN) loss, and $\mathcal{L}_{\text{cls}}$ and $\mathcal{L}_{\text{loc}}$ stand for the classification loss and bounding-box regression loss, respectively. The parameter $\lambda$ represents a hyper-parameter that necessitates fine-tuning to achieve optimal model performance.

\subsection{Data Augmentation}
\begin{figure}[t]
    \centering
    \includegraphics[width=\columnwidth]{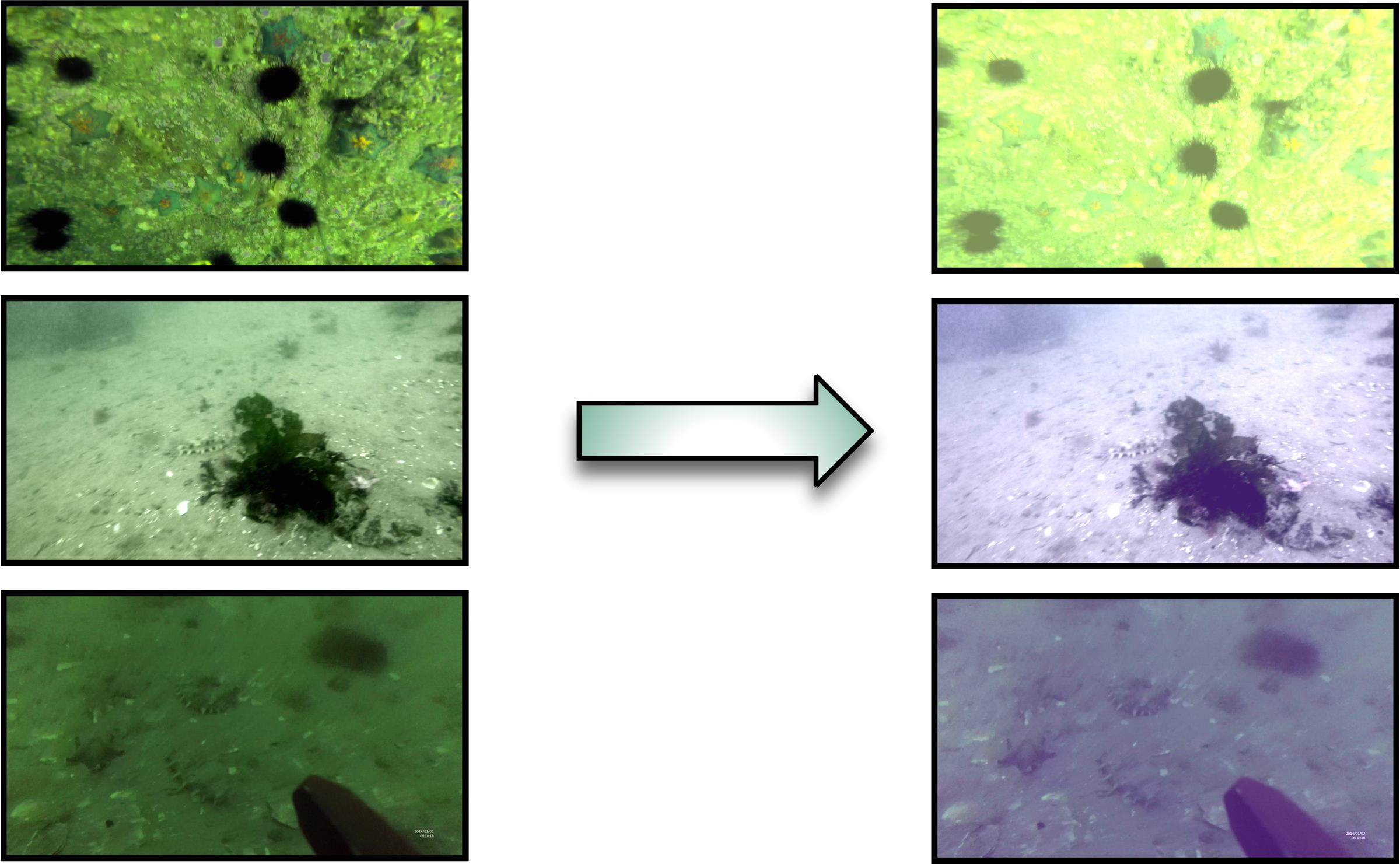}
    \caption{SCG* is a method that generates variations of each image in the dataset by solely altering the low-frequency information while preserving the core content but introducing different styles.}
    \label{SCG}
\end{figure}

Data augmentation helps alleviate domain shift by enriching the diversity of image styles. It can effectively expand the variety of styles in cases where the training set is either monotonous or limited in style. However, as the number of styles increases, manual labeling becomes more challenging, making it difficult to leverage DAL techniques. In contrast, with DSP, we could represent a greater number of domains using a smaller subset of domains. This implies that data augmentation methods could seamlessly integrate with DAL, even when dealing with a broader range of styles. To achieve this, we employed a Spurious Correlations Generator (SCG)\cite{xu2023multi} to generate a greater variety of stylistic images, see in fig \ref{SCG}. Specifically, SCG applied the  Discrete Cosine Transform (DCT) to the input images, transforming them into the frequency domain, and then blended them with randomly generated reference images in the frequency domain. We employed SCG* as our method, and what sets it apart from the original SCG in the research paper is that it exclusively manipulates the low-frequency information to obtain new styles, without any modifications to the high-frequency details. This was accomplished by adjusting the blending parameters to enhance stylistic diversity. As we acquired more domains, resulting in a more domain-invariant representation. This approach leverages the introduction of more stylistic diversity, thereby enhancing the model's adaptability to various domains and features. 

\begin{figure*}[htp]
    \centering
    \includegraphics[scale=1.0]{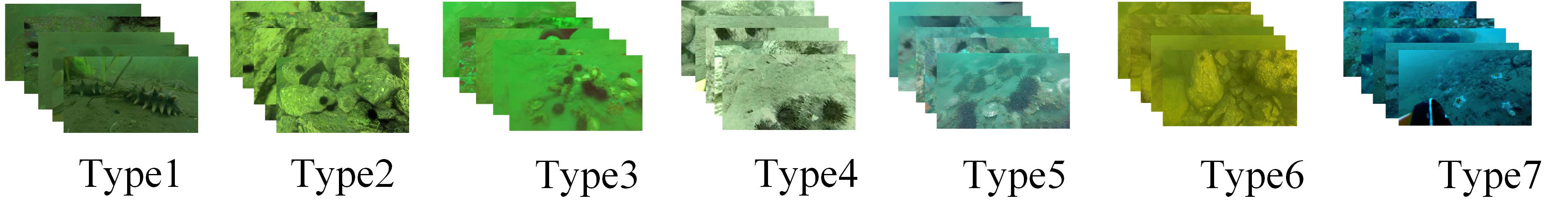}
    \caption{S-UODAC2020 is a dataset for underwater cross-domain object detection, comprising four marine species: echinus, holothurian, scallop, and starfish. The training set consists of 4,745 images sourced from six distinct domains, while the test set comprises 797 images from domains distinct from those in the training set.}
    \label{dataset}
\end{figure*}

\begin{table*}[ht]
\centering
\caption{The performance of various domain generalization methods on the benchmark S-UODAC2020, where 'Ave' represents mAP50}
\renewcommand{\arraystretch}{1.5}
\scalebox{0.9}{
\begin{tabular*}{\textwidth}{@{\extracolsep{\fill}}llp{1cm}llllll}

\hline
Method    & Backbone  & Epochs & Echinus(\%) & Starfish(\%) & Holothurian(\%) & Scallop(\%) & Ave.(\%) \\ \hline
DeepAll\cite{ren2015faster}   & ResNet50  & 12     & 74.79       & 36.59        & 43.12           & 40.94       & 48.86    \\
DG-YOLO\cite{liu2020towards}   & DarkNet53  & 12     & 62.74       & 26.83        & 32.84           & 34.54       & 39.24    \\
Mixup\cite{zhang2017mixup}     & ResNet50  & 12     & 70.23       & 34.58        & 40.01           & 18.86       & 40.92    \\
DANN\cite{ganin2015unsupervised}      & ResNet50  & 12     & \textbf{78.62}       & 42.76        & 50.60           & 43.48       & 53.87    \\
DANN\cite{ganin2015unsupervised}      & ResNet101 & 24     & 73.23       & 49.92        & 50.96           & 50.61       & 56.18    \\
CCSA\cite{motiian2017unified}      & ResNet50  & 12     & 76.71       & 36.85        & 40.58           & 37.46       & 47.90    \\
CrossGrad\cite{shankar2018generalizing} & ResNet50  & 12     & 77.67       & 45.43        & 49.80           & 42.40       & 53.83    \\
MMD-AAE\cite{li2018domain}   & ResNet50  & 12     & 75.73       & 35.00        & 43.31           & 44.86       & 49.73    \\
CIDDG\cite{li2018deep}     & ResNet50  & 12     & 76.37       & 39.89        & 42.27           & 43.65       & 50.55    \\
CIDDG\cite{li2018deep}     & ResNet101 & 24     & 74.04       & 48.98        & 49.71           & 45.67       & 54.60    \\
JiGEN\cite{carlucci2019domain}     & ResNet50  & 12     & 76.15       & 39.06        & 50.27           & 41.44       & 51.73    \\
JiGEN\cite{carlucci2019domain}     & ResNet101 & 24     & 75.92       & 47.01        & 51.37           & 46.50       & 55.20    \\
DANN+ELS\cite{zhang2023free}  & ResNet50  & 12     & 76.39          & 42.47           & 48.49              & 43.07          & 52.61       \\
DMCL\cite{chen2023achieving}      & ResNet50  & 12     & 78.44       & 54.62        & 53.15           & \text{59.23}       & 61.36    \\
RoIAttn\cite{liang2022excavating}      & ResNet50  & 12     & 74.41       & 43.28        & 50.01           & 42.66       & 52.59    \\
VFNet\cite{zhang2021varifocalnet}      & ResNet50  & 12     & 72.97       & 43.21        & 44.02           & 47.84       & 52.01    \\
\textbf{Ours}      & ResNet50  & 12     & 76.27          & \textbf{57.23}           & \textbf{53.59}              & \textbf{60.41}          & \textbf{61.88}   \\
\hline

\end{tabular*}
}
\label{table1}
\end{table*}

\section{EXPERIMENTAL RESULTS AND ANALYSIS}
\subsection{Experiments on the S-UODAC2020 benchmark}
\subsubsection{Experimental setup}
The experiments were conducted with NVIDIA GeForce RTX 3080Ti GPU. Following benchmark S-UODAC2020, the training set consists of six different domains, labeled as type1 to type6, while the evaluation and testing are conducted on the seventh domain, referred to as type7, as depicted in Fig \ref{dataset}. We have chosen the classic Faster R-CNN model as our detector, augmented with a Feature Pyramid Network (FPN) to enhance its detection capabilities. For the implementation of Faster R-CNN, we have used the mmdetection \cite{chen2019mmdetection} version 2.24.1 framework. The backbone chosen for this model is ResNet-50. We set the batch size to 1 and trained the model for 12 epochs. The optimizer used is SGD (Stochastic Gradient Descent) with a learning rate of 0.005, weight decay of 0.0001, and momentum of 0.9. We did not employ multi-scale training, and all images were resized to a consistent size of (1333, 800) pixels during training. Only the horizontal-flip data augmentation method is employed unless specified otherwise. We extract 64-channel feature maps from the backbone for feature statistics. For the training of the domain classifier in the DSP module, we opt for the utilization of 128 base domains. We iterate through 800 training iterations, then inferring the domain classifier for each image, ultimately obtaining a 128-dimensional domain label for each image. As for the trade-off parameter $\lambda$ in domain adversarial learning, we have selected a value of 0.7.
\subsubsection{Comparison with other domain generalization methods}
The comparison of domain generalization methods is shown in Table \ref{table1}. Faster R-CNN + FPN approach exhibits limited generalization capabilities, while Mixup, a commonly employed data augmentation technique, has demonstrated limited effectiveness in underwater scenarios and can even lead to adverse effects. In contrast, DANN showcases exceptional performance, holding significant advantages over other methods. However because of the constraints posed by discrete, our approach consistently outperforms them, outperforms them even with a ResNet101 backbone. In the S-UODAC2020 benchmark, our method outperforms all others, establishing itself as state-of-the-art in this benchmark.


\begin{table}[t!]\centering
\caption{The performance of various types of domain labels, including one-hot encoding, Environment Label Smoothing (ELS), and labels obtained under various DSP training epochs.}
\renewcommand{\arraystretch}{1.5}
\begin{tabular}{cccc}
\hline
Labeling Method & mAP  &  Labeling Method & mAP\\
\hline
DeepAll & 48.86 &  DANN(one-hot) & 53.87 \\
DANN(ELS) & 52.61 & DANN(DSP\_100) &54.21 \\
DANN(DSP\_500) & 51.67 & DANN(DSP\_800) & 54.50\\
DANN(DSP\_1000) & 51.40 & DANN(DSP\_4500) &52.31 \\
\hline
\end{tabular}
\label{table6}
\end{table}

\subsection{Experiments on the Cityscapes benchmark}
We utilized the Cityscapes \cite{wu2022single} for our experiments. The training dataset consisted of 19,395 daytime-sunny images sourced from BDD100K \cite{yu2020bdd100k}. Our test dataset comprised 26,158 night-sunny images from BDD100K, along with an additional 3,775 images collected from the Foggy Cityscapes \cite{sakaridis2018semantic} and Adverse-Weather \cite{hassaballah2020vehicle} datasets. This approach allowed us to assess whether our method, trained primarily on readily available data, could perform effectively under more challenging conditions.

We used the results obtained with \cite{wu2022single} as our baseline. To align with its methodology, we employed the Faster R-CNN+FPN object detection network, with ResNet-101 serving as the backbone detector. The image size was set to have a minimum side length of 600 pixels. Our Faster R-CNN implementation was based on mmdetection version 2.24, with a learning rate of 0.0025, weight decay set at 0.0001, and a momentum of 0.9. The hyperparameters for the DSP was kept consistent with those used in the S-UODAC2020 dataset. In total, our model underwent 10 epochs of training to achieve the final results. Due to variations in the implementation of Faster R-CNN, there may be differences in the values of FPN. In the table, we use "FPN*" to denote this. As we can observe in table \ref{nightsunny}, in the night-sunny environment, FPN outperforms all other domain generalization methods except for our own. Despite the differences in the values between our re-implemented FPN* and FPN, our approach still surpasses the best-performing methods. In the daytime-foggy scenario as in table \ref{foggy}, our method exhibits overwhelming superiority over the remaining methods.

\begin{table}[t!]
\caption{The performance of various domain generalization methods on the night-sunny scene.}
\renewcommand{\arraystretch}{1.5}
\resizebox{\columnwidth}{!}{
\begin{tabular}{ccccccccc}
\hline
Method   & bus  & bike & car  & motor & person & rider & truck & mAP  \\ 
\hline
FPN      & 37.4 & 33.1 & 62.2 & 21.4  & 42.5   & 32.1  & 40.9  & 38.6 \\ 
FPN*     & 41.6    & 38.2    & 67.3    & 20.9     & 50.3      & 32.0     & 46.3     & 42.4    \\
SW\cite{pan2019switchable}       & 35.4 & 28.6 & 56.7 & 18.4  & 38.2   & 26.2  & 39.3  & 34.7 \\
IBN-Net\cite{pan2018two}  & 40.2 & 31.4 & 62.1 & 19.0  & 42.9   & 29.3  & 44.2  & 38.4 \\
IterNorm\cite{huang2019iterative} & 28.8 & 29.2 & 55.7 & 12.3  & 35.9   & 25.4  & 35.4  & 31.8 \\
ISW\cite{choi2021robustnet}      & 37.4 & 32.2 & 60.4 & 16.5  & 41.0   & 29.2  & 43.0  & 37.1 \\
Ours     & \textbf{42.9}    & \textbf{39.9}    & \textbf{68.0}    & \textbf{23.6}     & \textbf{50.8}      & \textbf{32.4}     & \textbf{47.5}     & \textbf{43.6} \\
\hline   
\end{tabular}
\label{nightsunny}
}
\end{table}
\begin{figure*}[htp]
    \centering
    \includegraphics[scale=0.55]{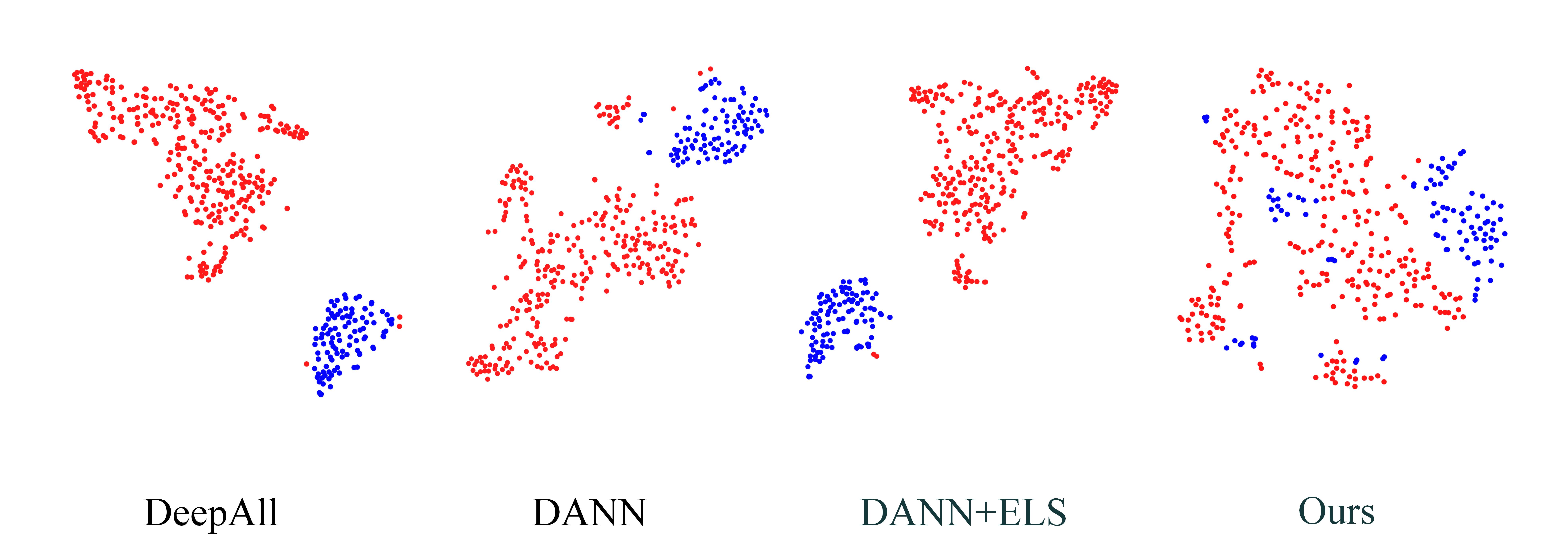}
    \caption{Visualizing different methods using t-SNE. Red points denote data from the source domain, while blue points represent data from the target domain. These features are extracted from the final stage of ResNet using various methods.}
    \label{tsne}
\end{figure*}
\begin{figure*}[htp]
    \centering
    \includegraphics[scale=0.24]{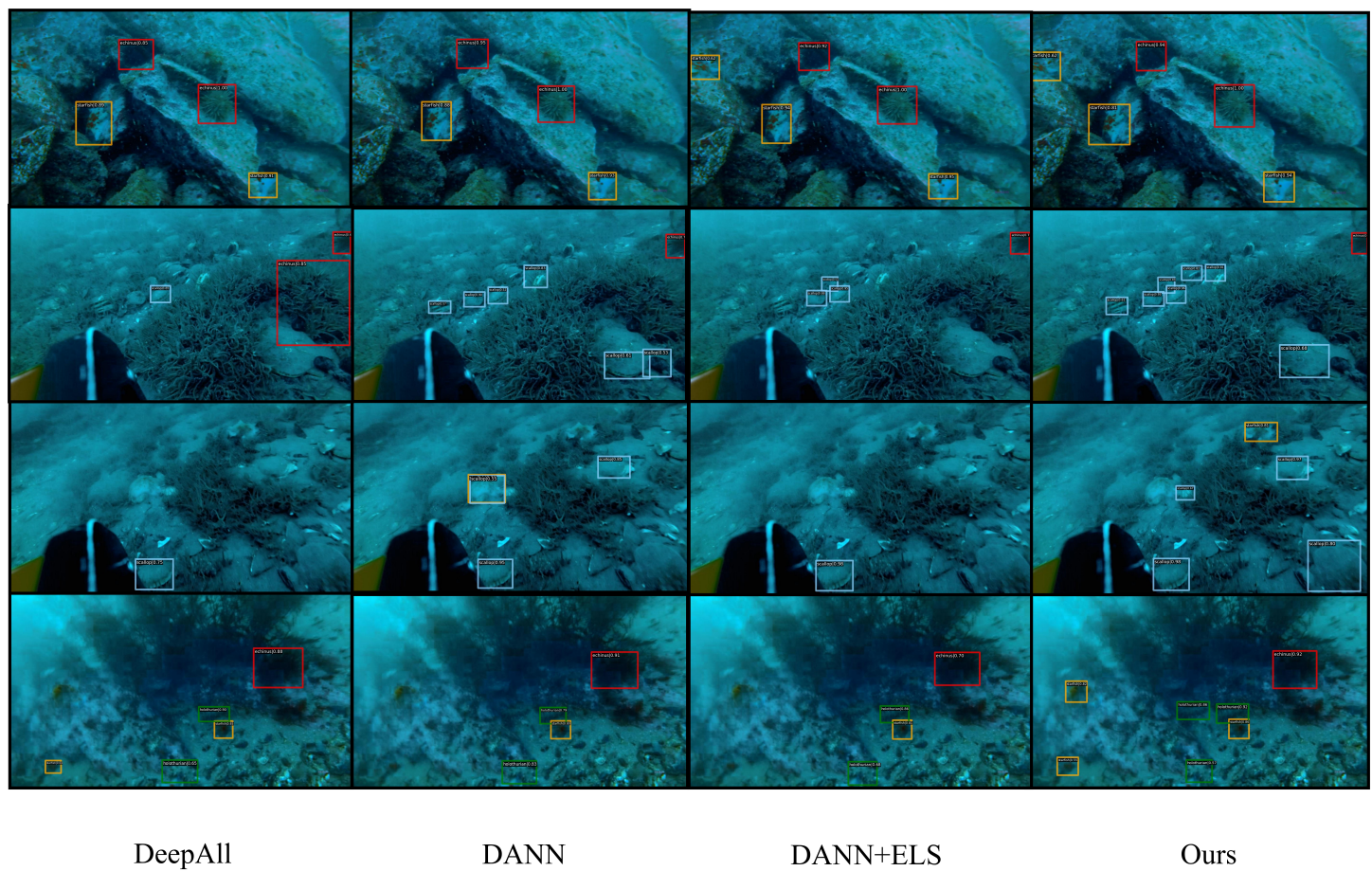}
    \caption{Comparison of actual detection results under different methods. Different colored boxes represent the discovery of various underwater creatures. The red box signifies the presence of echinus, the blue box represents scallop, the yellow box indicates starfish, and the green box denotes holothurian.}
    \label{visual}
\end{figure*}

\begin{table}[t!]
\caption{The performance of various domain generalization methods on the daytime-foggy scene.}
\renewcommand{\arraystretch}{1.5}
\resizebox{\columnwidth}{!}{
\begin{tabular}{ccccccccc}
\hline
Method   & bus  & bike & car  & motor & person & rider & truck & mAP  \\ \hline
FPN      & 30.5 & 29.7 & 52.1 & 28.4  & 33.9   & 40.4  & 21.0  & 33.7 \\
FPN*     & 29.9    & 31.3    & 57.6    & 30.5     & 36.5      & 40.7     & 21.0     & 35.4    \\
SW\cite{pan2019switchable}       & 32.0 & 28.4 & 52.3 & 28.8  & 33.5   & 39.5  & 21.9  & 33.8 \\
IBN-Net\cite{pan2018two}  & 32.5 & 31.4 & 52.5 & 31.1  & 38.0   & 42.1  & 23.5  & 35.9 \\
IterNorm\cite{huang2019iterative} & 25.3 & 27.4 & 50.4 & 24.0  & 32.2   & 37.4  & 18.6  & 30.7 \\
ISW\cite{choi2021robustnet}      & 31.9 & 30.5 & 51.9 & 30.8  & 37.5   & 40.9  & 21.9  & 35.1 \\
Ours     & \textbf{35.6}    & \textbf{33.6}    & \textbf{61.3}    & \textbf{35.2}     & \textbf{39.3}      & \textbf{44.4}     & \textbf{24.0}     & \textbf{39.1}    \\ \hline
\end{tabular}
\label{foggy}
}
\end{table}
\begin{figure*}[t]
    \centering
    \begin{minipage}{\textwidth}
        \centering
        \includegraphics[width=0.48\textwidth]{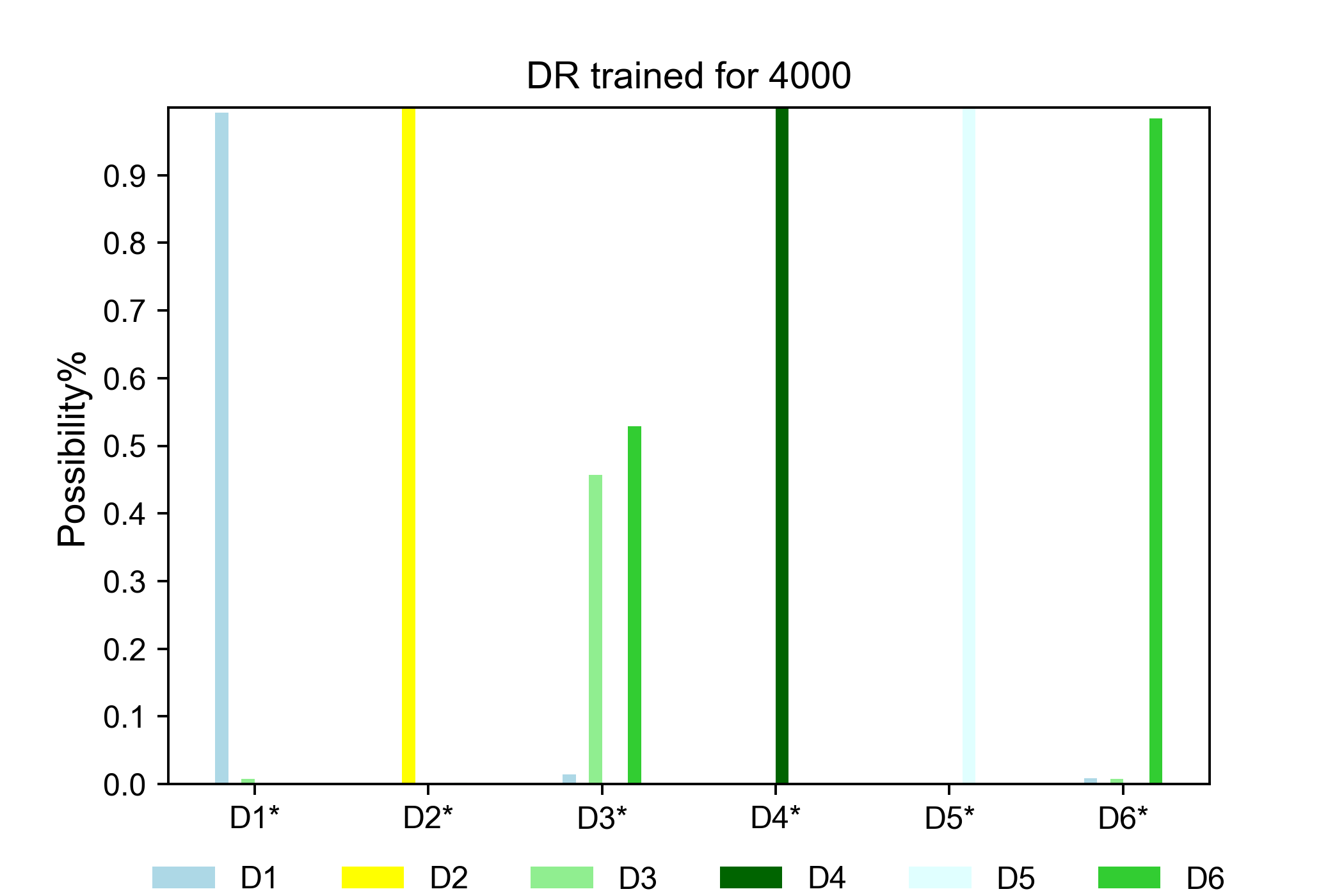}\hfill
        \includegraphics[width=0.48\textwidth]{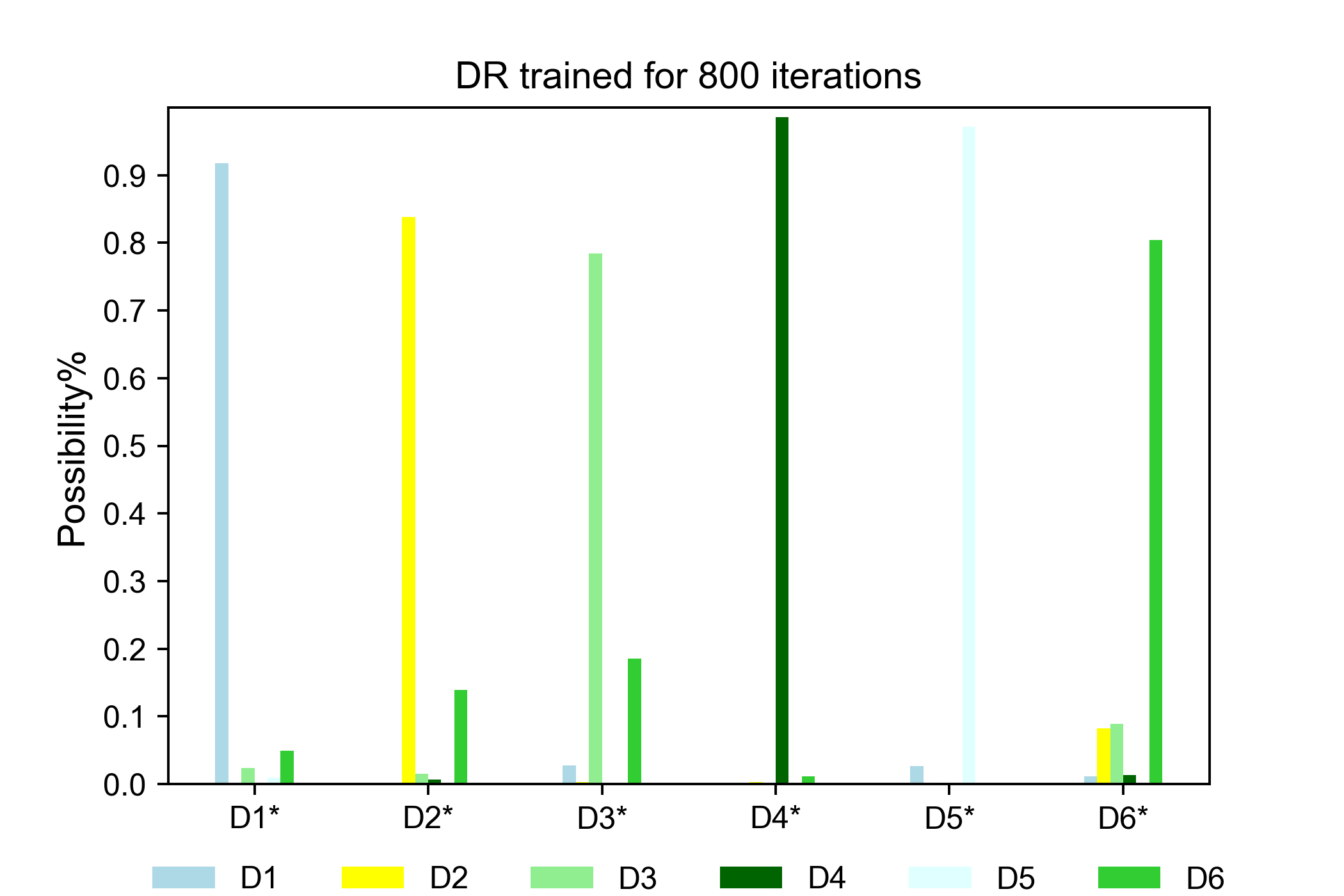}
    \end{minipage}
    \caption{$D_i^*$ denotes an image intentionally labeled as the i-th domain by humans, while $D_i$ represents the i-th base domain selected by DSP. $D_i^*$ is considered as a probability combination of $D_1$ to $D_6$. (left) Pseudo domain labels obtained after 4000 iterations of DSP training. (right) Pseudo domain labels obtained after 800 iterations of DSP training.}
    \label{statistic}
\end{figure*}

\begin{figure}[!t]
    \centering
    \resizebox{\columnwidth}{!}{
    \includegraphics[scale=0.55]{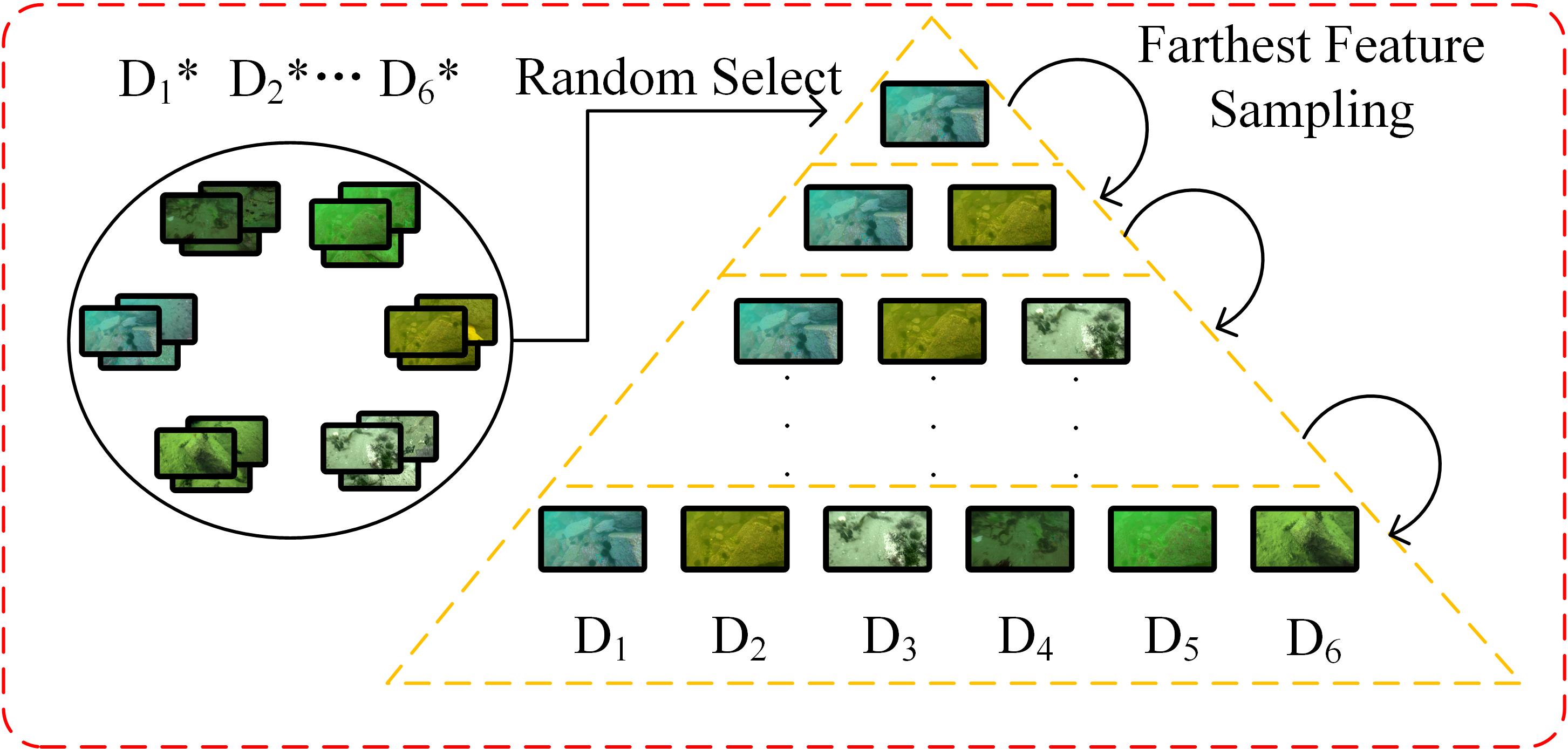}}
    \caption{When the number of base domains equals the manually partitioned domain count, utilizing Farthest Feature Sampling can identify domains $D_1$ through $D_6$ within the source domains $D_1^*, D_2^*, ... D_6^*$.}
    \label{pyramid}
\end{figure}


\begin{table}[t]
\centering
\caption{Ablation study of the number of the base domains $K$ in S-UODAC2020 dataset. $K=0$ denotes DeepAll.}
\renewcommand{\arraystretch}{1.5}
\begin{tabular}{lccccc}
\hline
$K$ & 0 & 2 & 3 & 4 & 5 \\
\hline
mAP & 48.9 & 51.9 & 53.0 & 52.2 & 52.7 \\
\hline
\end{tabular}
\label{table9}
\end{table}

\begin{table}[t]
\caption{Ablation study of SCG* across different datasets. $K$ is set to 128.}
\renewcommand{\arraystretch}{1.5}
\resizebox{\columnwidth}{!}{
\begin{tabular}{lcccc}
\hline
\multicolumn{3}{l}{\textbf{I. S-UODAC2020}}
 & &  \\
\textbf{Method} & \textbf{DeepAll} & \textbf{DSP w/o SCG*} & \textbf{SCG*-only} & \textbf{DSP + SCG*} \\
mAP & 48.9 & 54.5 & 58.8 & 61.9 \\
\hline
\multicolumn{3}{l}{\textbf{II. Sunny $\rightarrow$ Foggy}}
 & & \\
\textbf{Method} & \textbf{DeepAll} & \textbf{DSP w/o SCG*} & \textbf{SCG*-only} & \textbf{DSP + SCG*} \\
mAP & 35.4 & 35.3 & 37.7 & 39.1 \\
\hline
\multicolumn{3}{l}{\textbf{III. Sunny $\rightarrow$ Night}}
 & & \\
\textbf{Method} & \textbf{DeepAll} & \textbf{DSP w/o SCG*} & \textbf{SCG*-only} & \textbf{DSP + SCG*} \\
mAP & 42.4 & 42.1 & 42.2 & 43.6 \\
\hline
\end{tabular}
\label{table5}
}
\end{table}

\subsection{Ablation studies}
\subsubsection{Pesudo labels analysis}
When the number of the base domains selected matches the artificially partitioned domains in the dataset as shown in Fig \ref{pyramid}, the base domains chosen by DSP align with the manually designated domains. For example, the selected $D_1$ resembles $D_1*$.
As illustrated in Fig \ref{statistic}, longer training of DSP results in an over-confident label prediction, while shorter training of DSP can smooth the label and capture more relations between different domains.
Table \ref{table6} studies the training iterations of DSP.
The optimal performance is achieved when DSP reaches 800 iterations. A remarkable finding is that, even without the use of any manual annotations, the pseudo labels generated by DSP surpass the performance of manually annotated one-hot labels and even outperform the results obtained by softening the manual annotations using ELS. This model underscores the capability of DSP to provide more accurate domain labels for domain adversarial training.

\subsubsection{The number of the base domains}
In the previous section, we set the base domains to exactly match the manually partitioned domains. We further tested scenarios with fewer or more domains selected by DSP. As demonstrated in the last row of TABLE \ref{table9}, even when the number of base domains is reduced to 2, the proposed method still outperforms the baseline DeepAll ($K=0$). The domain diversity in S-UODAC dataset is limited so we only set $K$ less than the numbers of source domains. 

\subsubsection{Combining domain data augmentation with DSP}
Domain data augmentation can largely enrich the domain diversity in the dataset, DSP can leverage that to further improve the performance. We employed SCG* in different datasets to generate various styles of images. From the results in Table \ref{table5}, using SCG* along can improve the performance because it enriches the training data. 
Using DSP alone can improve the performance in S-UODAC2020, while DSP provides no improvement in both Sunny $\rightarrow$ Foggy and Sunny $\rightarrow$ Night scenarios of the Cityscapes dataset because they are single-source domain generalization problems. 
This result shows the dependence of DSP on the domain diversity in the dataset. If we combine SCG* with DSP, we can further improve the performance in all three datasets because DSP effectively excavates the invariant features from diverse training domains.

\subsubsection{T-SNE visualization}
T-SNE visualization was employed to observe feature distributions after applying different methods, which is shown in Fig \ref{tsne}. We extracted features from the final layer of ResNet50, selecting 300 random images from the source domain and 100 from the target domain for visualization. In the graph, blue dots represent the target domain, while red dots represent the source domain. As depicted in the visualization, our approach brings the feature distances between the two domains closer, establishing a connection between the tasks of detecting the source and target domains. Consequently, this approach yields the best results.
\subsubsection{Visualization of detection results} We present the most visually compelling results as shown in Fig \ref{visual}. Both vanilla Faster R-CNN and DANN exhibit false positive detections when recognizing objects in the second image, whereas ELS experiences a significant number of false negatives. In contrast, our method not only identifies a greater number of target objects but also avoids any false positive detections. Consequently, if our model is utilized for underwater exploration in unfamiliar environments, it has the potential to identify a greater number of aquatic organisms present in the water.

\subsection{CONCLUSION}
This paper aims to address challenges faced by domain adversarial training in underwater scenes, where over-confident discrete manual domain labels lead to the instability of adversarial training.
We propose the Domain Similarity-Perceived Label Assignment (DSP), representing each image based on its similarity to a set of base domains. The proposed approach demonstrates outstanding performance on the S-UODAC2020 and Cityscape datasets. The results suggest that smooth and continuous label space can effectively improve the performance of domain adversarial training. We believe that the applicability of DSP extends beyond this, as it can be employed in various directions such as cross-domain pedestrian re-identification.

\bibliographystyle{IEEEtran}
\bibliography{Mybib}
 
\end{document}